\documentclass[journal]{IEEEtai}
\usepackage[utf8]{inputenc}
\usepackage{cite}
\usepackage[cmex10]{amsmath}
\usepackage{amsmath}
\usepackage{amsfonts}
\usepackage{amsthm}
\usepackage{algorithm}
\usepackage{algorithmic}
\usepackage{multirow}
\usepackage{multicol}
\usepackage{booktabs}
\usepackage{amsmath}
\usepackage{algorithm}
\usepackage{multicol,multienum}
\usepackage{multirow}
\usepackage[caption=false,font=footnotesize]{subfig}
\usepackage{color}
\usepackage{graphicx}
\usepackage{xspace}
\usepackage{tabularx}
\usepackage{array}
\newcolumntype{P}[1]{>{\centering\arraybackslash}p{#1}}

\begin{document}

\title{A Transistor Operations Model for Deep Learning Energy Consumption Scaling Law}

\author{Chen~Li,~Antonios~Tsourdos~and~Weisi Guo,~$\IEEEmembership{Senior~Member,~IEEE}$ 
\thanks{

Chen Li, Antonios Tsourdos and Weisi Guo are with Digital Aviation Research Technology Centre (DARTeC), Cranfield University, Bedford, United Kingdom. Weisi Guo is also with the Alan Turing Institute, London, United Kingdom. Corresponding author: weisi.guo@cranfield.ac.uk.
}}

\maketitle

\begin{abstract}

Deep Learning (DL) has transformed the automation of a wide range of industries and finds increasing ubiquity in society. The high complexity of DL models and its widespread adoption has led to global energy consumption doubling every 3-4 months. Currently, the relationship between the DL model configuration and energy consumption is not well established. At a general computational energy model level, there is both strong dependency to both the hardware architecture (e.g. generic processors with different configuration of inner components- CPU and GPU, programmable integrated circuits - FPGA), as well as different interacting energy consumption aspects (e.g., data movement, calculation, control). At the DL model level, we need to translate non-linear activation functions and its interaction with data into calculation tasks. Current methods mainly linearize nonlinear DL models to approximate its theoretical FLOPs and MACs as a proxy for energy consumption. Yet, this is inaccurate (est. 93\% accuracy) due to the highly nonlinear nature of many convolutional neural networks (CNNs) for example. 

In this paper, we develop a bottom-level Transistor Operations (TOs) method to expose the role of non-linear activation functions and neural network structure in energy consumption. We translate a range of feedforward and CNN models into ALU calculation tasks and then TO steps. This is then statistically linked to real energy consumption values via a regression model for different hardware configurations and data sets. We show that our proposed TOs method can achieve a 93.61\% - 99.51\% precision in predicting its energy consumption. 
\end{abstract}

\begin{IEEEImpStatement}
Machine learning is one of the fastest growth areas for computational resources (300,000x from 2012 to 2018, doubling every 3-4 months). Data centres are predicted to dominate over 20\% of global energy consumption by 2030. What is important is to understand the relationship between the deep neural network (DNN) model and energy consumption, even if it co-varies with hardware. In this paper, we develop a bottom-up Transistor Operations (TOs) method to expose the role of both (1) non-linear activation functions and (2) DNN structure in energy consumption. We show that our proposed TOs method can achieve a 93.61\% - 99.51\% precision in estimating energy consumption. The method can help developers predict energy consumption before training to 93.61\% - 99.51\% precision and trade-off model performance with sustainability.
\end{IEEEImpStatement}

\begin{IEEEkeywords}
Energy Consumption; Deep Learning; Model Architecture; Transistor Operations; 
\end{IEEEkeywords}

\section{Introduction}

\begin{figure*}
     \centering
     \includegraphics[width=1.0\linewidth]{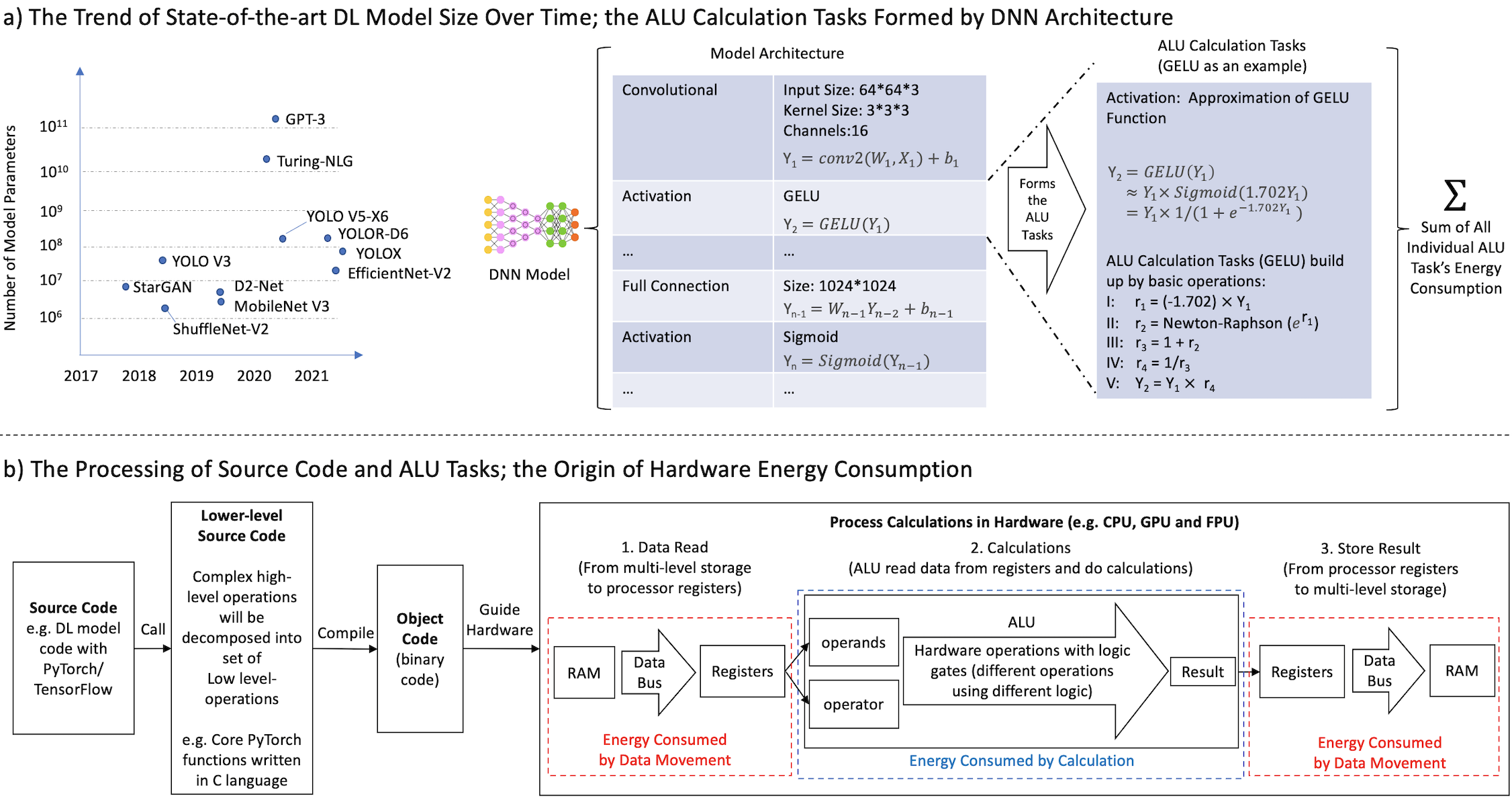}
     \caption{The Trend of DL Model Size Over Time and the Origin of DL Energy Consumption}
     \label{intro}
\end{figure*}

\IEEEPARstart{R}{apidly} increased AI demand has generated a huge increase in computational resource requirement (300,000x from 2012 to 2018) \cite{energyshift}. Energy consumption in data centres around the world to maintain data and learn models will account for 10\% of global energy consumption in 2025 and 20.9\% in 2030 \cite{theinformation}. The green challenge of achieving sustainable and environmentally friendly computing in the data age is gaining increasing attention from industry and academia \cite{bigdata}. The endless chasing of higher-precision in deep learning spawns ultra-large-scale models, especially in computer vision (CV) and natural language processing (NLP) \cite{energyNLP,DLenergy} (see - Fig.~\ref{intro}\textit{a}: YOLOR-D6 for CV with 174.7 million parameters in May 2021; openAI GPT-3 for NLP with 175 billion parameters in May 2020). Researchers in \cite{greenai} argue that this trend is unfriendly to computational resources, energy and the global environment. Developers should carefully analyze the requirements (e.g. precision, robustness) and backgrounds (e.g. computational hardware, energy supply) of DL tasks to make a trade-off between the performance and economy of DL models. The network size of large-scale DL models also barriers their deployment in energy-sensitive devices (e.g. Drones and remote sensors). Developers urgently need a theoretical method to analyze the scaling law of DL model energy consumption during model configuration changes to design and select energy-efficient DL models.

Indeed, there are now widespread efforts to reduce the size of neural network architectures both post-training \cite{ANG}, federated to the edge \cite{federated}, and more recently during training \cite{guo}. We will discuss these in detail later.

\subsection{Review on Deep Learning Energy Model}

DL energy consumption in training, validation, and testing can all be related to the model complexity, data size, and hardware implementation \cite{DLhardware}. As shown in Fig.~\ref{intro}\textit{a}, the DL model architecture (incl. the activation function) determines the resulting execution order of the equivalent arithmetic logic units (ALU) tasks. Within this, the input data affects the DL model hyper-parameters which is part of the ALU tasks and overall they all contribute to different energy consumption. Based on the information above, any comprehensive analysis of DL energy model must encompass: 
\begin{itemize}
    \item analyze the origin of hardware energy consumption at transistor operation (TO) and data movement level
    \item Translate linear and non-linear DL activation functions into the real calculation tasks executed by ALU
    \item Propose a TO based energy metric that is generalised across diverse hardware operations
    \item Develop a regression model to quantify the energy scaling with model configuration
\end{itemize}

Fig.~\ref{intro}\textit{b} (left to right) briefly demonstrates the running of DL model code in hardware and resulting energy consumption. We provide a review of different processors and their energy consumption, please refer to more details in \textbf{Appendix A}. Once the DL model source code (e.g. Python) is run, learning frameworks (e.g. PyTorch, TensorFlow) will call the relevant core functions written in lower-level languages (e.g. C/C++). For example, the GELU \cite{GELU} operation will be decomposed into a set of ALU-supported instructions (see - Fig.~\ref{intro}\textit{a}). Lower-level source codes will be further compiled into object codes to guide data reading, calculation and result storage \cite{machinecode}. The calculation in ALU follows a designed process \cite{ALU}. Firstly, operands will be called from data storage (DRAM etc.) to the processor registers through data bus. The energy for moving one unit data varies with the memory hierarchy. Secondly, do the specific hardware operation (specified in operator) on operands, and generate calculation energy consumption. Thirdly, the result will be temporarily stored in registers and then moved to other levels of storage. We provide a review of DL energy optimization, if interested, please refer to more details in \textbf{Appendix B}.

\subsection{Deep Learning's Energy Breakdown}

Current literature shows that DL models primarily consume two types of energy: (1) calculation energy (as described earlier), and (2) data movement and memory access energy. The latter constitutes a significant part of the energy consumption, especially for large data sets \cite{memoryDL, MITmethod, MITPruning}. In the execution of each instruction, data calling from hierarchy storage dominate up to \textit{90\%} energy consumption while calculations in ALU count less than \textit{30\%} \cite{MITeyeriss}. 

\begin{figure}
     \centering
     \includegraphics[width=1.0\linewidth]{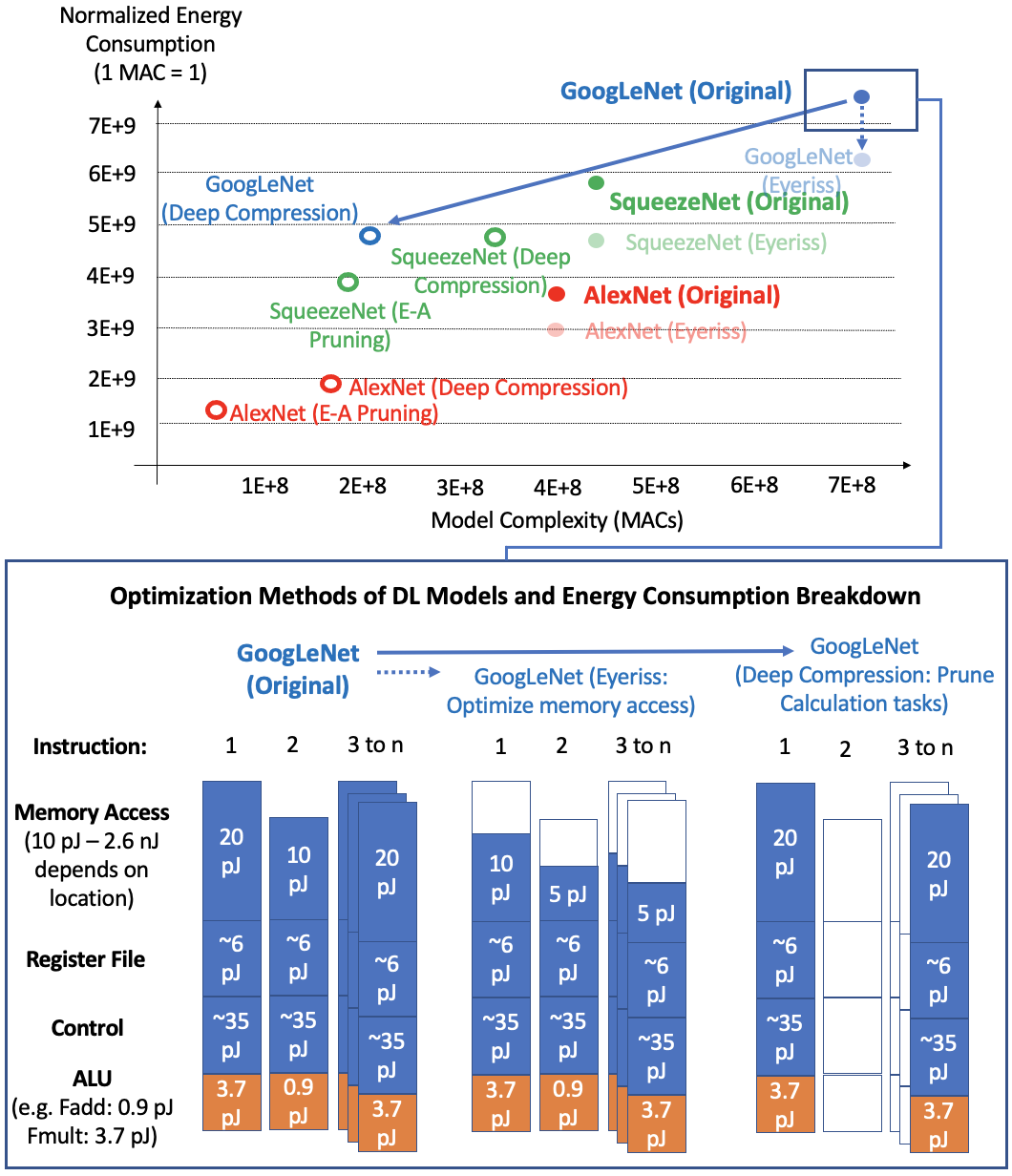}
     \caption{(top) Normalized Energy Consumption and (bottom) Energy Breakdown of DL Models. Data from \cite{breakdown,MITPruning})}
     \label{MACsEnergy}
\end{figure}

As shown in Fig.~\ref{MACsEnergy}(bottom), the data to be moved depends on operation tasks in instructions. Data movement optimization methods like Eyeriss \cite{MITeyeriss} and MONeT \cite{memoryopt} significantly reduce memory access energy consumption with higher data multiplexing rate and low-cost storage usage. Network pruning and compression methods (e.g. energy-aware pruning \cite{eapruning} and Deep compression \cite{deepcompression}) directly cut down calculation tasks to slash overall energy consumption. This means the investigation of DL calculation energy cost is still beneficial in energy-aware network light-weighting (e.g. using binary neural networks and doing network pruning) in energy-sensitive DL scenarios, optimizing neuron structure (e.g. efficient activation functions) and designing energy-efficient application-specific integrated circuits (e.g. TPU) for DL. The maximum energy efficiency of deep learning models can only be achieved through combined optimization of both computation and data movement energy consumption.

\begin{table*}[ht]
\centering
\begin{tabular}{P{80pt}P{35pt}|P{28pt}P{35pt}P{33pt}|P{35pt}P{50pt}|P{25pt}P{85pt}}
		\toprule
        Summary & Precision (\%) & Theory Basis & Experiment Data & Focus Area & Data Movement Energy & Calculation Energy & Energy Metric & Disadvantages\\ \midrule
        
         Experimentation of System Level Energy Consumption Increase \cite{SIMD} 
         & 63.05-73.7 
         & $\times$  
         & \checkmark
         & SIMD \& No. Data Access
         & \checkmark
         & \checkmark ~(No distinction of operation types)
         & Joule
         & No analysis of the relationship between the DL model configuration and energy consumption\\\midrule
        
        Experimentation Layer-based DL Energy Consumption and Core Usage \cite{layer-energy} 
        & 97.21 (avg) 
        & $\times$  
        & \checkmark
        & Total Run-time Power 
        & \checkmark
        & \checkmark ~(No distinction of operation types)
        & Joule
        & Not distinguishing linear and non-linear operations in layer configuration and energy consumption, need information about hardware run time and power\\ \midrule
        
        Theoretical Analysis of DL Complexity based on Only Linear Operations \cite{FLOPs}  
        & 92.9-99.5 
        & \checkmark 
        &  $\times$ 
        & Calculation FLOPs 
        & $\times$ 
        & \checkmark ~(Linear Operations)
        & FLOPs
        & Ignored Non-linear Operations (e.g., activation functions)\\ \midrule
        
        Theoretical \& Experimental Analysis of DL Energy Consumption based on Data Movements \cite{MITmethod}    
        & 96.4-96.5 
        & \checkmark 
        & \checkmark 
        & Data Movement \& MACs
        & \checkmark 
        & \checkmark ~(Linear Operations)
        & MACs
        & Data sparsity is hard to estimate accurately; ignored non-linear operations
        \\\midrule
        
       Proposed Transistor Operations (TOs) Method in This Paper       
       & 93.6-99.5 
       & \checkmark 
       & \checkmark 
       & Transistor Level Operations 
       & $\times$
       & \checkmark ~(Linear and non-linear operations)
       & TOs/Joule
       & No data movement energy analysis
       \\ \bottomrule
\end{tabular}
\caption{Methods for DL Energy Consumption Estimation. Acronyms in Table: Deep Learning (DL), Single Instruction Multiple Data (SIMD), Transistor Operations (TOs), Floating-point Operations (FLOPs), Multiply–accumulate Operations (MACs).}
\label{comparison}
\end{table*}

\subsection{Related Energy Quantification Research}

There are several works investigating the energy consumption quantification and estimation of machine learning (ML) and deep learning (DL) models. In \cite{energyML}, the authors make a survey on machine learning energy consumption estimation approaches that use simulated hardware or performance monitoring counters (PMC). They find processor plays the main role rather then DRAM in the energy consumption of tree-based models with experimentation. As an extended work, in \cite{energyreview}, hardware energy observation tools and state-of-the-art machine learning energy consumption estimation approaches are reviewed and classified according to different techniques they use. Authors in \cite{IEEEenergyconsumption} proposed a lightweight code-level energy estimation framework for software applications with limited additional cost in resources and energy. However, it is a general method focused on achieving accurate energy observation of computational hardware without the perspective of ML/DL.

\textit{Synergy}, a method proposed in \cite{SIMD} uses linear regression on both the number of SIMD instructions and bus accesses observed from hardware PMC to measure and predict the energy consumption of CNNs at a layer-level. But no theoretical analysis of the relationship between CNN layer configuration and energy consumption is given. \textit{NeuralPower} proposed in \cite{layer-energy} uses sparse polynomial regression method to model the power and runtime according to layer configuration parameters (e.g., batch size, kernel size, etc.) of key CNN layers, then applies the model to unseen layers for energy consumption estimation. However, the layer-level analyze can not distinguish linear and non linear operations in layer configuration. Theoretical analysis of the relationship between CNN layer configuration and energy consumption is not given. Floating-point operations (FLOPs) is used as a simple model-structure-based DL computational complexity measurement in \cite{FLOPs} without mapping to DL energy consumption. The calculation of FLOPs only consider linear floating-point operations (multiplication and accumulation) in full-connection/convolutional layers without non-linear operations and other layers (e.g. non-linear operations in AFs and operations in pooling layers). These ignored parts count a non-negligible part of DL model complexity (around \textit{5\%} in CNN model and \textit{15\%} in DNN model depends on the model configuration measured by our theoretical TOs model). Authors in \cite{MITmethod} propose a theoretical DL energy estimation method that uses the simulation of data movements between multiple layers of storage to quantify and normalize DL energy consumption with the energy for doing one MAC operation as the unit. However, their method ignore non-linear operations and the difference between hardware (energy for doing one MAC operating varies in different computation hardware, the rate of energy consumption for storage and computation varies with different hardware combinations).  No practical energy data is given, and the memory simulation method is not open to readers. As an extension work, they propose \textit{Eyeriss} in \cite{MITeyeriss}, which optimizes the data flow of CNN models using higher data reuse rates and less data movement from expensive storage. Authors in \cite{2021energy} review the parameter size, FLOPs and performance metric of several benchmark DL models. They find the computation and energy efficiency of hardware is affect by the precision of floating-point (FP) numbers (e.g. FP16, FP32), and propose that multiplicative factors take the majority responsibility of DL energy consumption. However, the work is observation-based without study the energy scaling law led by DL model configurations. 

The comparison between aforementioned DL energy estimation methods and our proposed TOs model are summarised in Table \ref{comparison}, which also list the individual precision of each method (declared in the original paper or measured by us).

\begin{figure*}
     \centering
     \includegraphics[width=1.0\linewidth]{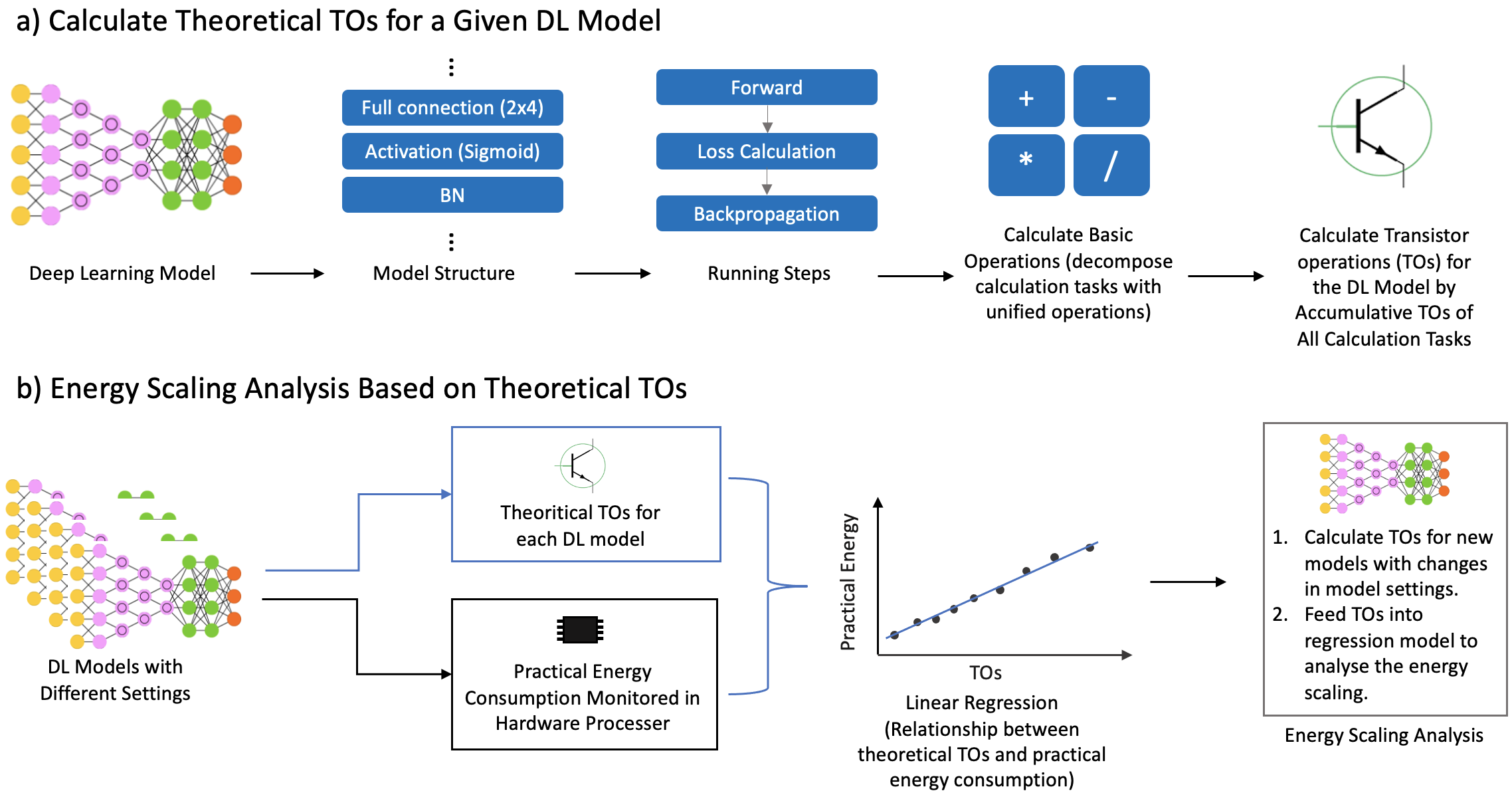}
     \caption{Flowchart for Calculating Theoretical TOs and Energy Scaling Analysis}
     \label{TOs}
\end{figure*}

\subsection{Gap Summary \& Innovation}

Currently, as we have seen from the above review, the relationship between the DL model configuration and energy consumption is not well established. At a general computational energy model level, there is both strong dependency to both the hardware architecture (e.g., CPU, GPU, FPGA), as well as different interacting energy consumption aspects (e.g., data movement, calculation, control). At the DL model level, we need to translate non-linear activation functions and its interaction with data into calculation tasks. 

As a summary, current methods mainly linearize nonlinear DL models to approximate its theoretical FLOPs and MACs as a proxy for energy consumption. Yet, this is inaccurate (est. 93\% accuracy) due to the highly nonlinear nature of many convolutional neural networks (CNNs) for example. 

In this paper, we develop an innovative bottom-level Transistor Operations (TOs) method to expose the role of nonlinear activation functions and neural network structure in energy consumption. We translate a range of feedforward and CNN models into ALU calculation tasks (e.g., basic operations (BOs)). Based on our TOs model, we demonstrate how the calculation energy scales when changing the model structure and activation functions. We also provide a verification experiment that compares the energy consumption estimated by TOs and the practical energy consumption monitored from general-purpose commercial processors (CPU, GPU). 

Compared with energy consumption estimation methods in \cite{energyreview,layer-energy}, our TOs model can individually analyze the calculation energy consumption with DL model structures, and consider non-linear operations that not included in \cite{FLOPs,MITmethod,layer-energy}. We will show that our proposed TOs method can achieve a superior 93.61\% - 99.51\% precision in estimating its energy consumption.

\subsection{TOs and the Scaling of DL Calculation Energy with DL Model Configuration}
The running of DL models involves extensive linear and non-linear (LNL) operations \cite{LNLDL}. These operations are processed in ALU by different processing logic which involves different numbers of transistors according to their individual factors (e.g. operation type) \cite{computerarchitecture}. The individual transistor operation requirements for each hardware operation (the total number of transistors used to complete the specified operation) provide a generic representation of TO. In this paper, we measure the energy consumption of DL models in terms of TOs, and propose a theoretical TOs model to analyze the scaling of DL energy consumption.

Due to the fact that hardware products are designed with different underlying processing logic (e.g. ripple-carry adder and carry look-ahead adder use different amount of transistors), manufacturing technology (e.g. semiconductor manufacturing process) and materials, the practical energy of running the same DL model on different hardware varies. The aforementioned factors certainly affect the energy consumption of each individual operation type, but only generate slight effects on the energy scaling relationship among different operation types. Indeed, analysing TOs individually on the design detail of each hardware will boost the accuracy of TO quantification. In this paper, our TOs model calculates the theoretical TOs for each hardware operation with generic processing logic, and we show it outperforms the FLOPs-based model in estimating the energy consumption of deep learning models in different hardware platforms.

\section{Theoretical Transistor Operations (TOs) Model}

The calculation of theoretical TOs for a given DL model is demonstrated as Fig.~\ref{TOs}$a$. Suppose the dataset have already been pre-processed (energy for data pre-processing is not considered in this paper, but TOs could be extended to data processing). Firstly, the layer list will be extracted from the model structure and settings (e.g. 2$\times$4 full-connection layer, activation layer with Sigmoid AF). Secondly, extract running steps according to different analysis levels: \textit{training level} involves model forward, loss calculation and backpropagation (BP); \textit{validation level} involves forward and loss calculation; \textit{inference level} only focus on model forward. The reason is that each step has individual calculation logic and resulting different energy consumption, the calculation of theoretical TOs for different analysis levels will be based on the summary of step-level TOs (e.g. validation level TOs contains model forward TOs and loss calculation TOs). Thirdly, layer-wise analysing of how many BOs are needed for each running step based on their calculation logic. We use addition, subtraction, multiplication, division and root operations as the categories of BOs, for the reason that higher-level calculations (e.g. activation of neuron with Sigmoid as AF function - see Function \ref{Feedforward}) are assembled by these five BOs at the software level. Finally, we analyze how many transistor operations are theoretically involved in each basic operation processed by ALU calculation logic, and calculate the theoretical layer-based TOs based on the number of five BOs and data types (e.g. FP-16, FP-32).

\subsection{layer based Basic Operations (BOs)}

The BOs of a DL model are calculated according to the processing logic of each layer, related to Fig.~\ref{TOs}$a$: \textit{Calculate basic operations}, details are shown in Alg.~\ref{ALG_BOs}. Suppose the layer list extracted from model structure is $L = \{l_{1},l_2...l_i...l_{\text{output}}\}$. For each layer $l_i \in L$, the BOs calculator $C_{\text{f}}(l_i)$ and $C_{\text{bp}}(l_i)$ calculates and the number of five BOs needed for layer forward and backpropagation respectively, return the result as a list (format: $[n_{\text{add}},n_{\text{sub}},n_{\text{mul}},n_{\text{div}},n_{\text{root}}]$, $n$ means the number of each basic operation). $C_{\text{loss}}(l_{\text{output}})$ takes only the output layer $l_{\text{output}}$ as input to calculate BOs for loss calculation (one of the running steps). The algorithm will return: layer based forward BOs in $B_{\text{f}}$; layer based BP BOs in $B_{\text{bp}}$; and BOs for loss calculation $b_{\text{loss}}$. We provide two examples that calculate the forward BOs for a full-connection layer and a convolutional layer respectively as follows. The calculation of BOs for loss and model backpropagation is similar to model forward but follows their individual calculation logic.

\begin{equation}
\begin{split}
&\text{Potential of the \textit{j}-th Neuron:}\\
&\xi_{j} = b_j + \sum_{i=1}^{I} (w_{i,j}\times x_{i})\\
&\text{Activation (Sigmoid) of the \textit{j}-th Neuron :}\\
&y_j = S(\xi_{j}) = 1/1+e^{-\xi_{j}}
\end{split}
\label{Feedforward}
\end{equation}

The calculation logic for full connection layers is proposed in \cite{feedforward}. Suppose full-connection layer $l_\text{Full-connection}$ have $I$ inputs; $O$ outputs; use Sigmoid \cite{sigmoid} as AF, the calculation of output $y_j$ on input $x$, weight $w$ and bias $b$ could be demonstrated in equation \ref{Feedforward}. Here, $\xi$ represent neuron output before activated by AF, and $S$ is Sigmoid function. The BOs for forward on one data instance could be calculated by equation \ref{BO_F}:

\begin{align}
\begin{split}
 &C_{\text{f}}(l_\text{Full-connection}) = BOs_{\text{Potential}} + BOs_{\text{Activation}}\\
 &BOs_{\text{Potential}} = O\times I(\text{mul} + \text{add})\\
 &BOs_{\text{Activation}} = O\times (\text{sub}+\text{add}+\text{div}+\text{root})\\
 &C_{\text{f}}(l_\text{Full-connection}) = [(I+1)O,O,IO,O,O]
\end{split}
\label{BO_F}
\end{align}

Here, the counting of BOs is analyzed on the logic in equations. For example, the calculation of $\xi_j$ in equation \ref{Feedforward} involves an accumulation of multiplication results, and an addition operation for adding bias. The multiplication operation happens $I$ times; addition operation by accumulation repeat $I-1$ times; add bias need one additional addition operation; the total operations are $I$ multiplications and $(I-1)+1$ additions. To calculate $\xi$ will repeat the process above $O$ times, so the BOs for linear operations are $O\times I$ multiplications and additions. In equation \ref{ALG_BOs}, `\text{mul + add}' means 1 multiplication operation and 1 addition operation. The items in the return list from $C_{\text{f}}(l_\text{Full-connection})$ means the number of each basic operation needed for this layer (the order is detailed before).

\begin{figure}
     \centering
     \includegraphics[width=1.0\linewidth]{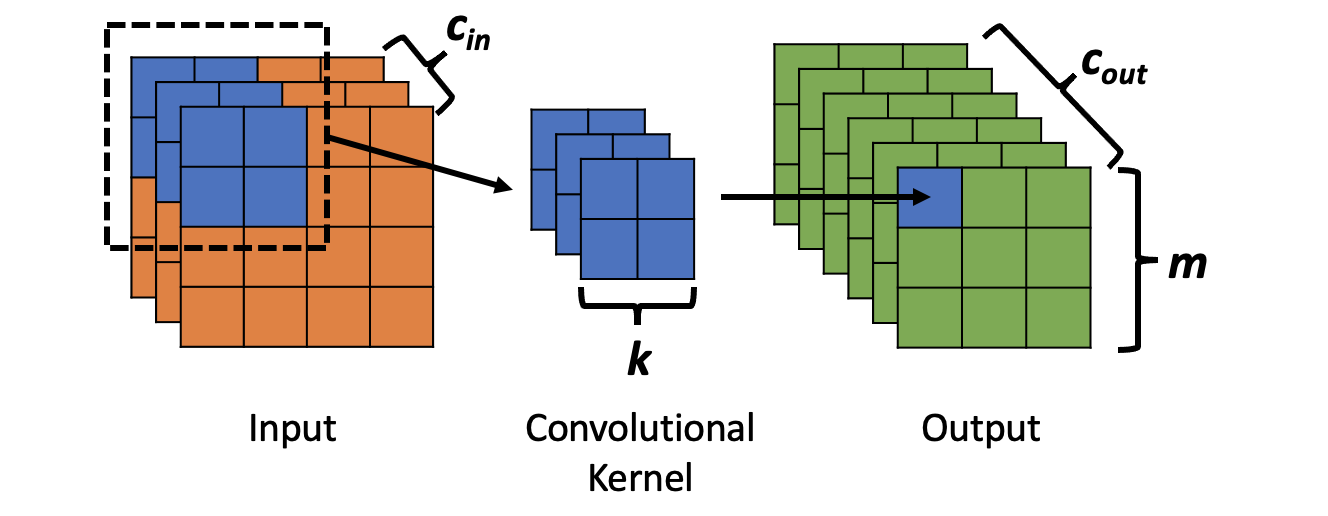}
     \caption{Convolutional Layer}
     \label{C}
\end{figure}

The working process of convolutional layers could be seen as Fig.~\ref{C}. Suppose $m,k,c_{\text{in}},c_{\text{out}}$ are the output window width, convolutional kernel size, number of input channels and number of output channels in convolutional layer $l_\text{Convolutional}$. With GELU (approximate as $\text{GELU}(x) = 0.5 \times x \times (1 + \text{Tanh} (sqrt(2 / \pi) \times (x + 0.044715 \times {x^{3}})))$ \cite{GELU} in \text{PyTorch}) as AF, the convolutional layer forward BOs on one instance could be calculated as equation \ref{BO_C}:

\begin{align}
\label{BO_C}
\begin{split}
 &C_{\text{f}}(l_\text{Convolutional}) = BOs_{\text{Convolution}} + BOs_{\text{Activation}}\\
 &BOs_{\text{Convolution}} = m^2\times c_{\text{out}}\times c_{\text{in}}\times k^2(\text{mul}+\text{add})\\
 &BOs_{\text{Activation}} = m^2c_{\text{out}}(\text{sub}+\text{add}+\text{div}+\text{root}+2\times \text{mul})\\
 &n_{\text{add}} = m^2c_{\text{out}}(1+c_{\text{in}}k^2)\\
 &n_{\text{sub}} = m^2c_{\text{out}}\\
 &n_{\text{mul}} = m^2c_{\text{out}}(2+c_{\text{in}}k^2)\\
 &n_{\text{div}} = m^2c_{\text{out}}\\
 &n_{\text{root}} = m^2c_{\text{out}}\\
 &C_{\text{f}}(l_\text{Convolutional}) = [n_{\text{add}},n_{\text{sub}},n_{\text{mul}},n_{\text{div}},n_{\text{root}}]
\end{split}
\end{align}

\subsection{Basic Operations, Data Type and Theoretical TOs}

In this step, the TO of DL model expressed by BOs will be further decomposed in transistor level to unify the energy metric for different basic operation types. As shown in Alg.~\ref{ALG_TOs}, TOs calculator $T(BOs,p)$ takes both the layer based BOs extracted by Alg.~\ref{ALG_BOs} and the data type $p$ used in DL model as input. $T$ opens the design logic of ALU, analyze the theoretical TOs for each operation modules (e.g. 32-bit adder) with their individual integrated circuit (IC) designs. And with the information of layer based BOs, $T$ is able to calculate the total theoretical TOs needed for DL model in different steps (e.g. validation). Function $T$ could be universally applied on the return lists of $C_{\text{f}}(l_i)$, $C_{\text{bp}}(l_i)$ and $C_{\text{bp}}(l_i)$ due to their similar data format (number of five basic operations). Alg.~\ref{ALG_TOs} will return layer based TOs for model forward $D_{\text{f}}$ and BP $D_{\text{bp}}$, as well as cumulative TOs for model forward $d_{\text{f\_total}}$, BP $d_{\text{bp\_total}}$, loss $d_{\text{loss}}$, and overall TOs $d_{\text{total}}$.

\subsubsection{BOs, Operation Units in ALU and Calculation of TOs}
In TOs calculator, the calculation of how many TOs are needed for a BO depends on the inner logic of ALU (open ALU and get number of logic gates, and open logic gates with transistors to count TOs). The various IC design for ALU inner logic follows the role of making a trade-off between time complexity (the time delay) and space complexity (number of transistors). The achieve of lower delay needs additional logic gates which cost more transistors and energy \cite{lowpoweradder} (e.g. adder: ripple-carry and carry look-ahead adder; multiplier: Wallace and Booth-Wallace multiplier). The logic of IC in ALU support different types of basic operations, however, not all the basic operations have their independent operation unit (e.g. adder is designed for addition, but subtraction is processed in adder by 2's complement). As multipliers are always built with adder as basic units, the changing of adder IC has a limited impact on the relationship between transistors in adder and multiplier, similar for other operation units. Although independent root operation units exist, they are not widely embedded in current PC processors. Root operation is always simulated with the Newton-Raphson method \cite{root} by operation units in ALU. TOs model in this paper uses the basic IC design to calculate the theoretical transistors needed by BOs. Please note, the design of IC is not focused in this paper, using the specific hardware IC may increase the analysis accuracy of model energy scaling on that hardware. Theoretically, transistors needed in adder follows the linear relationship to bit-length, multiplier and divider follow exponential relationship. Ten transistors are used for a NOR-XNOR gates based 1-bit full-adder as proposed in \cite{fulladder}. And according to the IC design in \cite{multiplieranddivider}, transistors used for the 64-bit Booth-Wallace multiplier and SRT divider are 90k and 110k respectively.

\begin{algorithm} [H]
	\caption{Calculate layer based BOs for DL model}
	\label{ALG_BOs} 
	\begin{algorithmic}
	    \REQUIRE DL model structure $L = \{l_1, l_2 ... l_{k}\}$
	    \REQUIRE Basic Operation Calculators: $C_{\text{f}}, C_{\text{bp}}, C_{\text{loss}}$
	    
		\STATE Initialise $B_{\text{f}} = [b_{\text{f}1}, b_{\text{f}2}, ..., b_{\text{f}k}]; B_{\text{bp}} = [b_{\text{bp}1}, b_{\text{bp}2}, ..., b_{\text{bp}k}]$
		\STATE Initialise $b_{\text{loss}} = [0,0,0,0,0]$ /* number of 5 BOs */
		\FOR{$l_i \in [1,2,...,k]$}
		    \STATE $b_{\text{f}i} = C_{\text{f}}(l_i)$
		    \STATE $b_{\text{bp}i} = C_{\text{bp}}(l_i)$
        \ENDFOR
        \STATE $b_{\text{loss}} = C_{\text{loss}}(l_{k})$
    \RETURN $B_{\text{f}}, B_{\text{bp}}, b_{\text{loss}}$
	\end{algorithmic} 
\end{algorithm}
\begin{algorithm} [H]
	\caption{Calculate Theoretical TOs Based on BOs}
	\label{ALG_TOs} 
	\begin{algorithmic}
	    \REQUIRE layer based BOs set $B_{\text{f}} = [b_{\text{f}1}, b_{\text{f}2}, ... ,b_{\text{f}k}]; B_{\text{bp}} = [b_{\text{bp}1}, b_{\text{bp}2}, ..., b_{\text{bp}k}]$
	    \REQUIRE BOs for loss calculation $b_{\text{loss}}$
	    \REQUIRE Data Type $p$
		\REQUIRE TOs calculator $T$
		\STATE Initialise $D_{\text{f}} = [d_{\text{f}1}, d_{\text{f}2}, ..., d_{\text{f}k}]; D_{\text{bp}} = [d_{\text{bp}1}, d_{\text{bp}2}, ..., d_{\text{bp}k}]$
		\STATE Initialise $d_{\text{f\_total}}, d_{\text{bp\_total}}, d_{\text{loss}}, d_{\text{total}} = 0,0,0,0$
		
		\FOR{$i \in [0,1,...,k]$}
		    \STATE $d_{\text{f}i} = T(b_{\text{f}i}, p)$
		    \STATE $d_{\text{bp}i} = T(b_{\text{bp}i}, p)$
		    \STATE $d_{\text{f\_total}} += T(b_{\text{f}i}, p)$
		    \STATE $d_{\text{bp\_total}} += T(b_{\text{bp}i}, p)$
		\ENDFOR
    \STATE $d_{\text{loss}} = T(b_{\text{loss}}, p)$
    \STATE $d_{\text{total}} = d_{\text{f\_total}}+d_{\text{bp\_total}}+d_{\text{loss}}$
    \RETURN $D_{\text{f}}, D_{\text{bp}}, d_{\text{f\_total}}, d_{\text{bp\_total}}, d_{\text{loss}}, d_{\text{total}}$
	\end{algorithmic} 
\end{algorithm}

\subsubsection{Floating-point Numbers in TOs Calculation}
\begin{figure}
     \centering
     \includegraphics[width=1.0\linewidth]{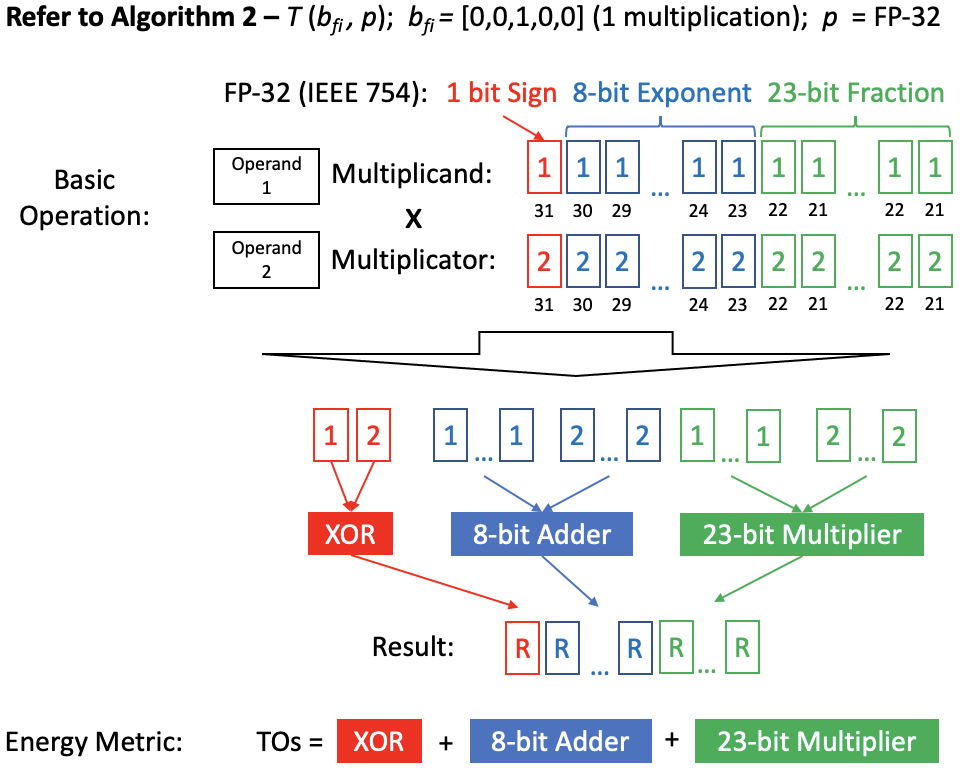}
     \caption{Calculation of Theoretical TOs based on BOs (Example: Multiplication Operation on Two IEEE 754 FP-32 Numbers)}
     \label{BOtoTO}
\end{figure}

Floating-point numbers (FP) are the most generally used data type in DL models. As FP used in majority programming language follows IEEE 754 standard \cite{floatpoint}, the calculation of theoretical TOs should consider the structure of binary FP with different precision (e.g. FP-32 known as single-precision floating-point, FP64 as double-precision floating-point) and calculation logic of FP numbers. According to IEEE 754, each FP-32 number contains an 1-bit sign, an 8-bit exponent and a 23-bit fraction (FP-64: 1-bit sign, 11-bit exponent and 52-bit fraction). Theoretically, adding two FP-32 numbers will need a 24-bit adder (23 full-adder and 1 half-adder without considering bit shift in the exponent). As shown in Fig.~\ref{BOtoTO}, the multiplication/division of two FP-32 numbers will need a XOR gate (for sign), a 24-bit multiplier/divider (fraction calculation) and an 8-bit adder (exponent calculation). It means the TOs for a FP-32 multiplication is theoretically be calculated by the sum of transistors for a 24-bit multiplier, a 8-bit adder and a XOR gate. As a theoretical model, transistors redundancy in practical hardware are not considered (e.g. multiplication of two 24-bit number on 32-bit multiplier).

\subsection{TOs Model and Energy Scaling}

Theoretical TOs for a given model could be calculated by Alg.~\ref{ALG_BOs} and Alg.~\ref{ALG_TOs}. The process of mapping the scaling of model TOs to its practical energy consumption scaling is demonstrated in sub-figure Fig.~\ref{TOs}$b$. 

Firstly, a list of DL models with different settings will be established. Secondly, calculate their individual TOs with our TOs model, and deploy them into hardware to collect their practical energy consumption separately. According to our theory, polynomial regression (PR) with different number of coefficients will be used to fit the relationship between practical energy consumption and TOs of DL models. To analyze the scaling of energy with a model design factor (e.g. width of hidden layers), a list of models will be generated by gradually changing the factor (e.g. models with 4,5,6 and 7 hidden layer width). Then, calculate their TOs and estimate their individual energy consumption by the previously fitted PR model. We demonstrate the energy scaling of a feed-forward DNN with different hidden layer widths and AFs in the next section, followed by that of a CNN model with different layer configurations. Please note that the practical energy consumption of models depends on different choices of hardware. If the hardware platform changes, the PR model should be retrained on energy data on the new hardware platform.

\begin{figure*}
     \centering
     \includegraphics[width=1.0\linewidth]{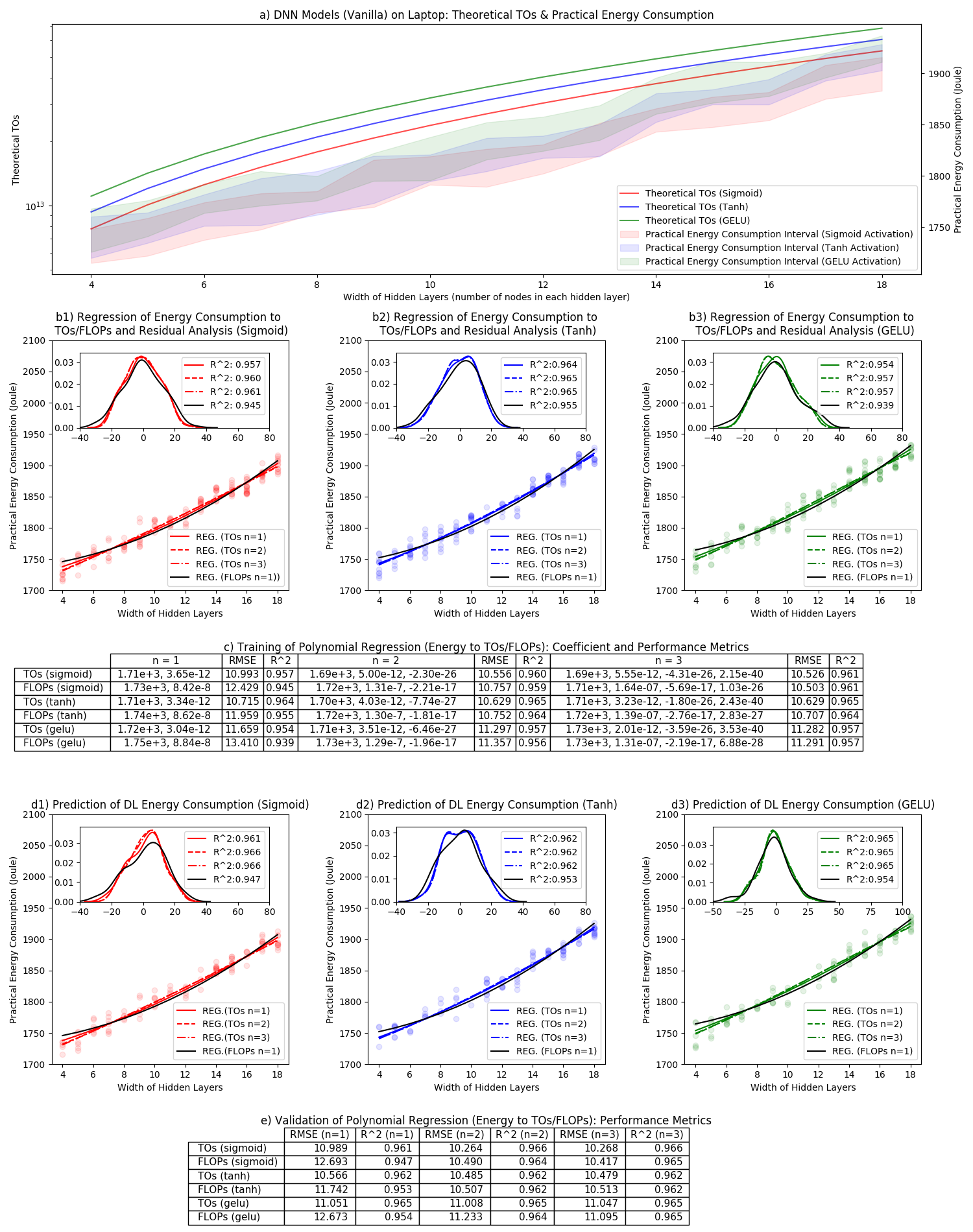}
     \caption{Method Verification (Vanilla Neural Network on Laptop CPU)}
     \label{verification_dnn_laptop_cpu}
\end{figure*}

\section{Method Verification}

To verify the theoretical TOs model, we design experiments to analyze the training energy scaling of 1) a feed-forward DNN set by changing the AFs and width of hidden layers; 2) a CNN set by changing the number of convolutional layers and kernel size in each layer. During the verification of energy scaling, the performance and structure rationality of each DL model will not be considered (Over-fitting will occur with over-large width). The experiment settings could be seen from Table \ref{exp}. 

\begin{table}[h]
\centering
\begin{tabular}{l|l|l}
\hline
Exp. Settings & DNN & CNN\\
\hline
Dataset & Banknote \cite{data}& Drone images  \\
Dataset Length & 1372 instances & 120 images\\
Data Structure & 4 inputs \& 1 output & 256*256*3 (RGB)\\\hline
Model Type & Feed-forward & Convolutional\\
Model Depth & 3 hidden layers & 2-10 C-layers\\
DNN Width & 4-18 nodes per layer & -\\
CNN Kernel (h\&w)& -& layer 2-6: 3\\
&-& layer 7-10: 4,5,6,7\\
CNN Channels& -& 32\\
Activation Functions & Sigmoid, Tanh, GELU & GELU\\
Batch Size & 64 & 64 (GPU: 16)\\

Epochs&2k (laptop CPU)& 25 (laptop CPU)\\
&5k (desktop CPU)& 20 (desktop CPU)\\
&& 100 (desktop GPU)\\

DL Framework & PyTorch& PyTorch\\
\hline\hline
Hardware & Laptop & Desktop\\
\hline
CPU & Intel Core i9 & AMD RYZEN 9\\
GPU & - & Nvidia 3080\\
OS & macOS Monterey& Windows 10\\
\hline
\end{tabular}
\caption{Experiment Settings and Hardware}
\label{exp}
\end{table}

The practical energy consumption of DL models is collected with \textit{Intel Power-Gadget software} \cite{Intelpower} (for Intel Architecture CPU\cite{intelcore}), \textit{OpenHardwareMonitor} (for AMD CPU) and \textit{TechPowerUp GPU-Z} (for GPU). These tools are designed using onboard energy sensors to collect the instantaneous power of processor cores, DRAM, and overall CPU/GPU package separately with timestamps \cite{Intelpower}. The energy consumption is calculated based on the integration of the instantaneous power consumption over time. Certain errors could exist in this value, as the challenge of energy monitoring is mentioned in \cite{energyML}. During the energy data collection, we force the program to run with a single CPU/GPU core. The reason is in the previous experiments we found data collected from multi-core shows huge differences across hardware platforms. That is because calculation communication and control costs among cores are overweight for small-scale calculation tasks, and significantly influence the processor power and program running time.

\subsection{Energy Consumption Scaling of Vanilla Neural Networks on Laptop CPU}

As shown in Fig.~\ref{verification_dnn_laptop_cpu}\textit{a}, we calculate TOs for feed-forward models with 4-18 width (number of nodes in each hidden layer) and use \textit{Sigmoid}, \textit{GELU} and \textit{Tanh}\cite{tanh} as AFs respectively. We did 50 experiments on each model setting with laptop CPU as introduced in Table \ref{exp}, drop the highest and lowest 5 data, and demonstrate the practical energy consumption interval of each model.
We further randomly split the energy data into training set ($60\%$) and validation set ($40\%$) for PR on DL model's practical energy consumption and TOs/FLOPs. We demonstrate the training of PR based on both FLOPs and TOs with 1-3 coefficients, and conduct a residual analysis in Fig.~\ref{verification_dnn_laptop_cpu}\textit{b1-b3} (we only demonstrate FLOPs-based PR with 1 coefficient in figures for higher figure clearness and readability, statistics for FLOPs-based PR could be seen in tables). In these figures, each point represents the practical energy consumption data collected from one experiment. We could see both FLOPS-based and TOs-based PR perform fairly excellent with their residuals following Gaussian distribution. However, as the R-square and RMSE shown in Fig.~\ref{verification_dnn_laptop_cpu}\textit{c}, TOs-based PR models outperform FLOPs-based PR models in fitting practical energy consumption of DL models with all AFs. But this superiority reduces with increasing the number of PR coefficients. We further demonstrate the validation of PR models in Fig.~\ref{verification_dnn_laptop_cpu}\textit{d1-d3}, and list the performance metric in Fig.~\ref{verification_dnn_laptop_cpu}\textit{e}. We also run the DNN set on a desktop CPU, please refer to Fig.~\ref{verification_dnn_desktop_cpu} in Appendix for your interest.

According to the coefficients shown in Fig.~\ref{verification_dnn_laptop_cpu}\textit{c} and Fig.~\ref{verification_dnn_desktop_cpu}\textit{c}, use data of Sigmoid with 1 coefficient (\textit{n=1}) for example, the relationship between DNN model's TOs and the practical energy consumption \textit{E} in current laptop CPU and desktop CPU could be demonstrated by function \ref{PRresult}.

\begin{equation}
\begin{split}
\text{DNN on laptop CPU:}\\
\text{E} = 1.71\times10^3 + 3.65\times10^{-12}\times \text{TOs}\\
\text{E} = 1.73\times10^3 + 8.42\times10^{-8}\times \text{FLOPs}\\
\text{DNN on desktop CPU:}\\
\text{E} = 9.96\times10^2 + 2.47\times10^{-12}\times \text{TOs}\\
\text{E} = 1.04\times10^3 + 5.71\times10^{-8}\times \text{FLOPs}\\
\end{split}
\label{PRresult}
\end{equation}

\subsection{Energy Consumption Scaling of CNN on Desktop GPU}

\begin{figure*}
     \centering
     \includegraphics[width=1.0\linewidth]{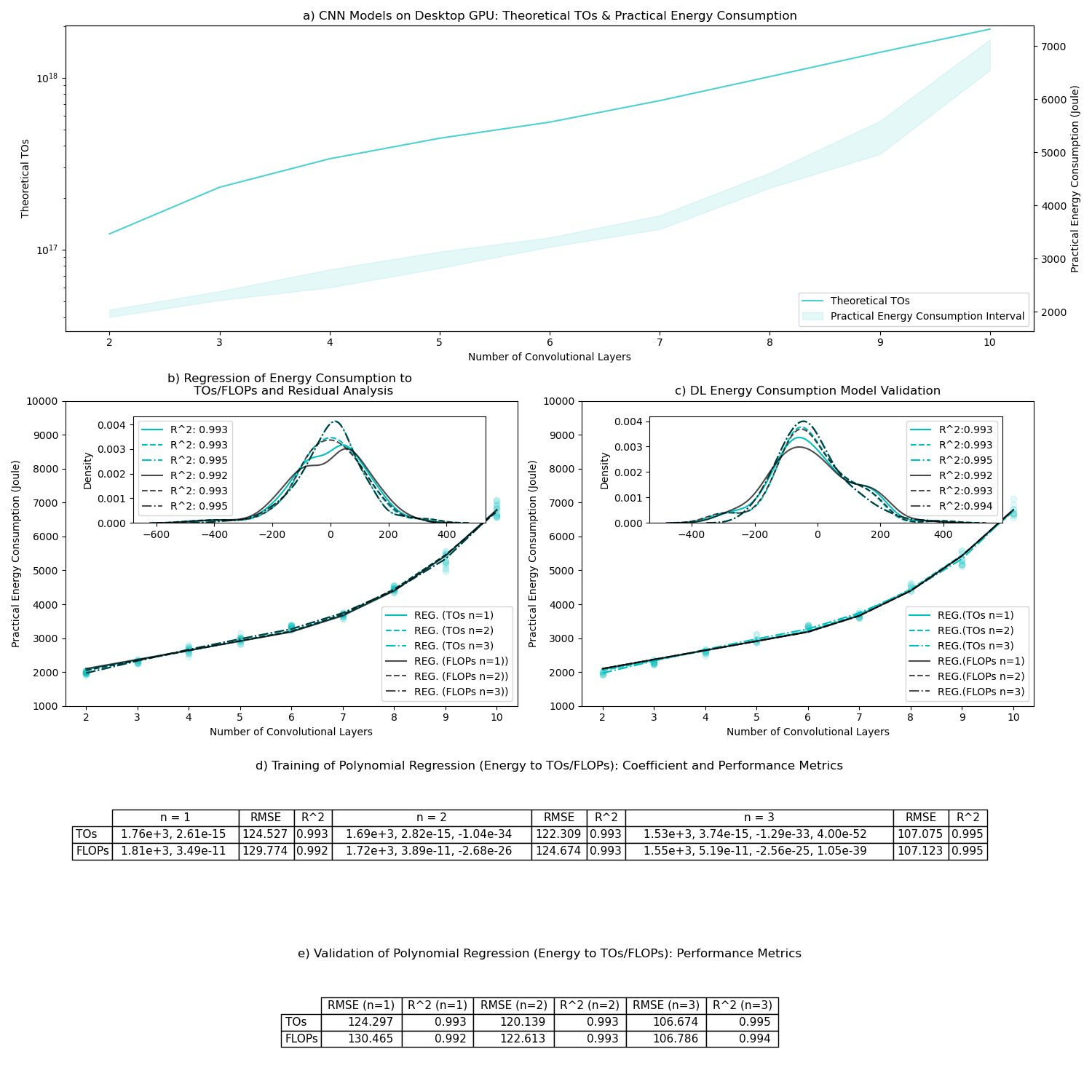}
     \caption{Method Verification (CNN on Desktop GPU)}
     \label{verification_cnn_desktop_gpu}
\end{figure*}

As shown in Fig.~\ref{verification_cnn_desktop_gpu}\textit{a}, we calculate TOs for CNN models with different depths (number of convolutional layers) and apply \textit{GELU} as the AF for each convolutional layer. The configuration of each CNN model could be seen from Table \ref{exp}. We did 50 experiments on each model, drop the highest and lowest 5 data, and demonstrate the practical energy consumption interval of each model.
The energy data is split randomly into training set and validation set with a rate of $60/40\%$. We demonstrate the training and validation of PR energy models with residual analysis in Fig.~\ref{verification_cnn_desktop_gpu}\textit{b-c}. In these figures, each point represents the practical energy consumption data collected from one experiment. We also summarise the PR model performance metric (R-square and RMSE) in Fig.~\ref{verification_cnn_desktop_gpu}\textit{d-e}. We also run the same CNN set on a desktop CPU and a laptop CPU separately, please refer to Fig.~\ref{verification_cnn_desktop_cpu} and Fig.~\ref{verification_cnn_laptop_cpu} in Appendix for your interest.

\begin{table}
\centering
\begin{tabular}{r|rr}
\hline
\multicolumn{3}{c}{\textbf{Desktop}}\\
\hline
\textbf{DNN: AFs} & \textbf{FLOPs} & \textbf{TOs}\\
\hline
Sigmoid (\%)&96.93-99.99 (avg 99.05)&97.46-99.99 (avg 99.14)\\
Tanh (\%)&96.81-99.97 (avg 99.04)&96.95-99.99 (avg 99.22)\\
GELU (\%)&97.11-99.97 (avg 98.94)&97.52-99.99 (avg 99.26)\\
Sigmoid (avg/max, J)&11.16/36.05 & 10.19/31.29\\
Tanh (avg/max, J)& 11.33/41.28 & 9.33/39.51\\
GELU (avg/max, J)& 12.78/35.55 & 9.04/29.45\\
\hline
\textbf{CNN: Processor} &&\\
\hline
CPU (\%) & 78.41-99.92 (avg 94.13) & 80.97-99.97 (avg 94.82) \\
GPU (\%)& 91.50-99.92 (avg 97.15) & 92.61-99.98 (avg 97.36)\\
CPU (avg/max, J)& 383.39/797.24& 341.46/747.77\\
GPU (avg/max, J)& 101.65/364.33& 95.95/362.66\\
\hline
\hline
\multicolumn{3}{c}{\textbf{Laptop}}\\
\hline
\textbf{DNN: AFs} & \textbf{FLOPs} & \textbf{TOs}\\
\hline
Sigmoid (\%)&98.14-99.99 (avg 99.41)&98.52-99.99 (avg 99.49)\\
Tanh (\%)& 97.95-99.98 (avg 99.45)&98.44-99.99 (avg 99.50)\\
GELU (\%)& 97.99-99.99 (avg 99.46)&98.40-99.99 (avg 99.51)\\
Sigmoid (avg/max, J)& 10.59/31.80& 9.31/25.56\\
Tanh (avg/max, J)&9.90/35.27&9.19/26.77\\
GELU (avg/max, J)& 9.85/34.57&9.19/30.47 \\
\hline
\textbf{CNN: Processor}&&\\
\hline
CPU (\%) & 74.03-99.90 (avg 92.93)& 76.45-99.99 (avg 93.61) \\
CPU (avg/max, J)& 501.09/1134.58& 453.25/1029.06\\
\hline
\end{tabular}
\caption{The Performance of FLOPs-based and TOs-based Method in DL Energy Estimation (Precision, Average Error and Max Error in Joule)}
\label{result}
\end{table}

According to the coefficients showed in Fig.~\ref{verification_cnn_desktop_gpu}\textit{c} and Fig.~\ref{verification_cnn_desktop_cpu}\textit{c}, the relationship (PR models with \textit{n=1}) between CNN model's TOs and the practical energy consumption \textit{E} in current desktop processors could be demonstrated by function \ref{PRresult_cnn}.

\begin{equation}
\begin{split}
\text{CNN on desktop GPU:}\\
\text{E} = 1.76\times10^3 + 2.61\times10^{-15}\times \text{TOs}\\
\text{E} = 1.81\times10^3 + 3.49\times10^{-11}\times \text{FLOPs}\\
\text{CNN on desktop CPU:}\\
\text{E} = 2.51\times10^3 + 4.29\times10^{-14}\times \text{TOs}\\
\text{E} = 2.67\times10^3 + 5.73\times10^{-10}\times \text{FLOPs}\\
\end{split}
\label{PRresult_cnn}
\end{equation} 

\begin{figure}
     \centering
     \includegraphics[width=1.0\linewidth]{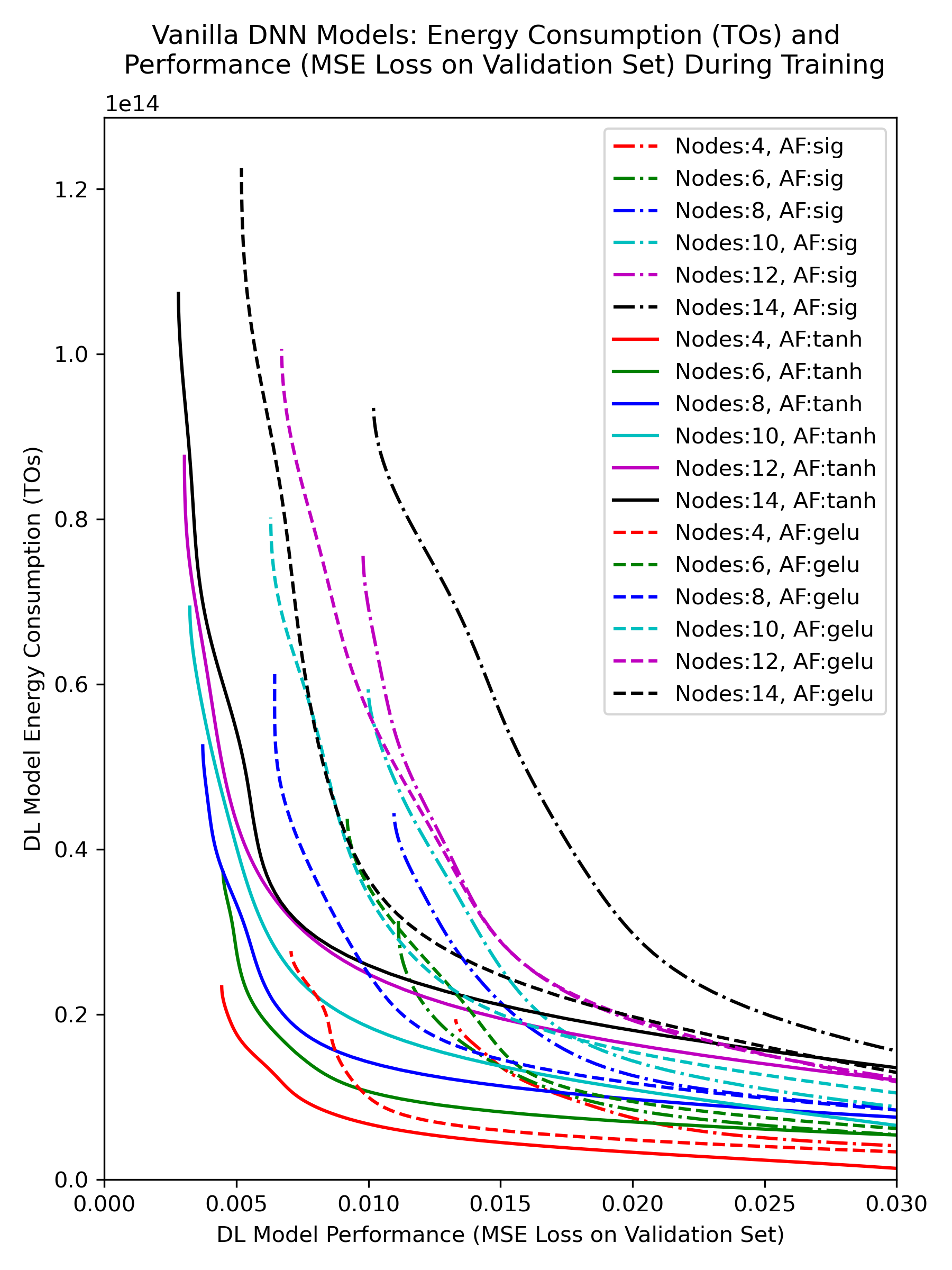}
     \caption{Energy Consumption (TOs) and Performance (MSE Loss on Validation Set) of Vanilla DNNs During Training}
     \label{TOs_Loss}
\end{figure}

Based on the experiments above, the performance of FLOPs-based and TOs-based PR models for DL energy estimation could be summarised by Table \ref{result}. From the table, TOs-based PR models perform better in all DL model configurations and hardware than FLOPs-based PR models. Compared with PR models on FLOPs, PR models on TOs achieve $0.14-2.56\%$ higher precision on DL energy consumption estimation, and a $10\%$ lower average estimation error in Joule.

We also show the energy consumption and the performance changes of DL models with different configurations during training (Vanilla DNN on desktop CPU)  in Fig.~\ref{TOs_Loss}, from where the energy efficiency of different DL models configurations (number of nodes and the choice of AFs) could be seen.

\section{Conclusion and Further Works}
Calculations form the core of deep learning training, which connects the DL model architecture and its energy consumption. In this paper, we proposed a TOs-based theoretical energy consumption model. Our TOs model consider both linear and non-linear DL operations and achieves a more accurate estimation of energy scaling compared with current FLOPs-based methods. We show that 1) theoretical TOs of DL shows a linear relationship with its real energy, and could be used as a metric for analysing the energy scaling of DL models; 2) TOs-based method outperforms FLOPs-based method with $0.14-2.56\%$ higher precision on DL energy consumption estimation, and lower 10\% of the average estimation error in Joule. Our proposed TOs could be extended to all DL models (e.g. RNN, Transformers) by the analysis of additional operations (e.g. Comparison Operator, automatic differentiation \cite{pytorch}). 

In future work, if we combined our proposed TOs method and the data movement energy quantification method \cite{MITmethod}, we can build a more holistic DL energy consumption framework and achieve a more accurate estimation of DL energy consumption.

\bibliographystyle{IEEEtran}
\bibliography{reference}

\appendix

\subsection{Review of Calculation Hardware - Data Storage and Processors}

 Mathematical operations involve two main types of hardware: multiple-levels of data storage (e.g. Disk, DRAM, Cache) and processors (e.g. CPU, GPU, TPU) for data storage/access and calculation respectively. Processors all use ALU as the fundamental calculation unit, with different configuration of inner components (e.g. ALU, control units) to achieve different features \cite{computerarchitecture}. An operation could be processed in different types of processor, and resulting different time cost and energy consumption. This means, when changing network configuration, the scaling law of calculation TO for different processor types is the same. However, the TO-determined energy consumption varies with the choice of processor type.

\subsection{Review of DL Energy optimization: Data Movement Energy and Calculation Energy}

Source code of a DL model could indirectly control the order and number of hardware operations, thereby determining the model energy consumption. Suppose the source code of a given learning model contains a set of calculation tasks $N=\{n_{1},n_{2}...n_{k}\}$. Task $n_i$ calls data $d_i$ from storage level $l_i$ (determined by data storing strategy), and calculate them with operation type $t_i$. The hardware-level energy consumption of a DL model is composed by calculation and data movement energy consumption\cite{MITmethod}. As for code running environment, the energy consumption level of the current data movement structure is represented by data movement energy metric $\theta_{m}$. Hardware calculation energy consumption level is represented by $\theta_{c}$ depending on hardware-related factors (e.g. circuit logic and materials). As function $f_{\text{move}}(d_i,l_i,\theta_{m})$ and $f_{\text{calculate}}(d_i,t_i,\theta_{c})$ calculates the energy consumption for data movement and calculation respectively, the energy consumption $E$ of doing one prediction with the given DL model could be summarised by equation \ref{energy}.

\begin{equation}
\begin{split}
    E = \sum_{i=1}^{k} (f_{\text{move}}(d_i,l_i, \theta_{m}) + f_{\text{calculate}}(d_i,t_i, \theta_{c}))
    \label{energy}
\end{split}
\end{equation}


\begin{equation}
\begin{split}
 \min_{m\in M} \text{trade-off} = \alpha E_m + (1-\alpha)loss_m
    \label{tradeoff}
\end{split}
\end{equation}

The analysis of data movement energy and calculation energy are in different significance for developers. As shown in equation \ref{energy}, the optimization of DL energy involves several factors. As data movement logic is pre-defined at the system level, it can only provide developers with very limited assistance in model designs. However, it helps the learning frame developers to optimize the data structure and move strategy to improve data movement efficiency (optimize $\theta_{m}$) as the method proposed in \cite{MITeyeriss}. By contrast, the analysis of calculation energy allows developers to intuitively understand the relationship between DL model architecture and operations happened in hardware. They can further optimize the learning model architecture (e.g. use energy efficient operation types, reduce number of tasks $k$ \cite{ANG}) to develop energy efficient learning models. At the same time, hardware engineers can design specialised hardware (e.g., TPU) for deep learning to increase the energy efficiency (optimize $\theta_{c}$) \cite{TPU}. Federated learning \cite{federated} with TPU in edge can make profit from both high energy efficiency and high learning efficiency \cite{TPUedge}. As shown in equation \ref{tradeoff}, developers could make a trade-off between DL model energy consumption and performance by a selected $\alpha$, and select the most appropriate model $m$ from candidate model set $M$.

\subsection{Extra Results of Method Verification}

\begin{figure*}
     \centering
     \includegraphics[width=1.0\linewidth]{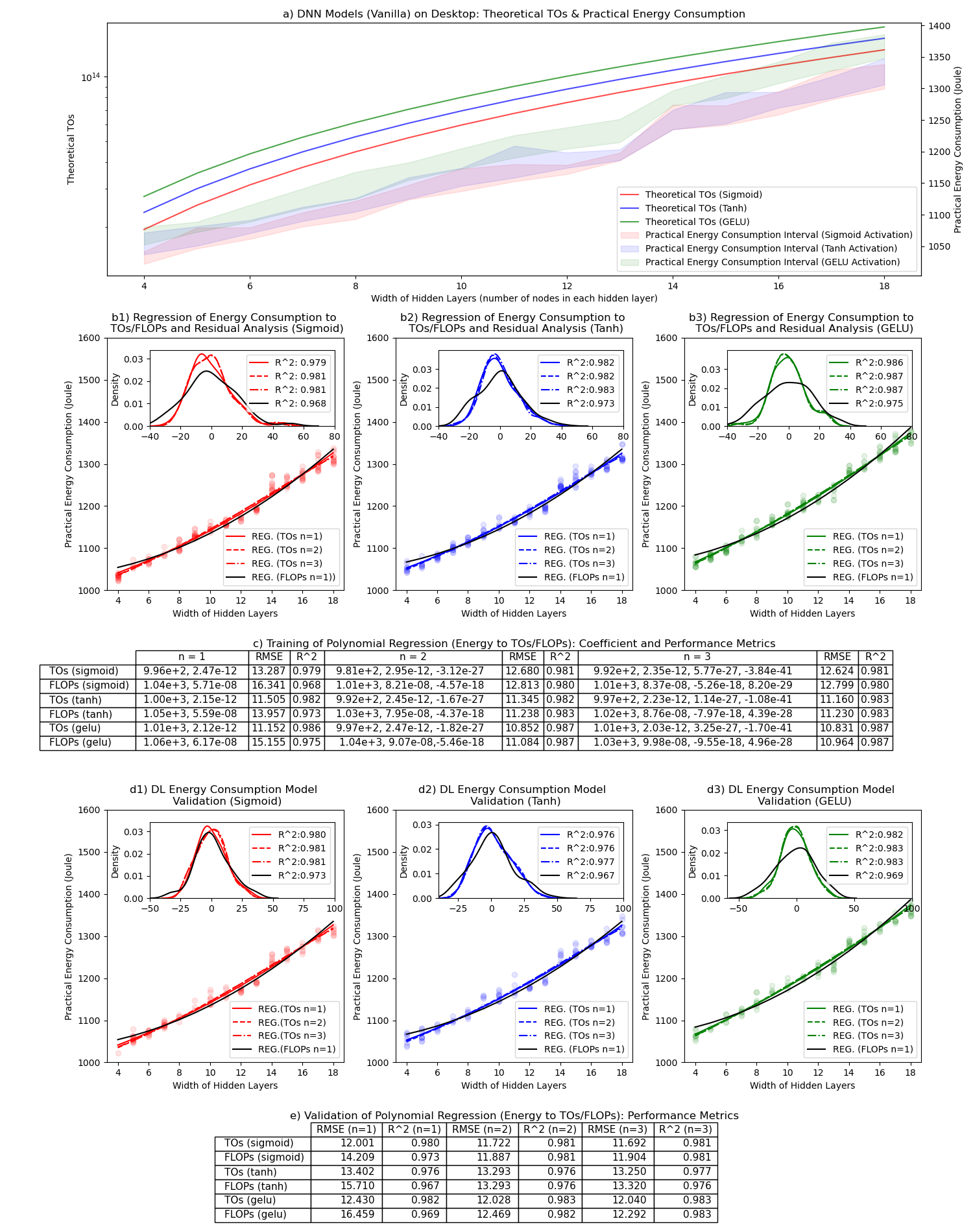}
     \caption{Method Verification (Vanilla Neural Network on Desktop CPU)}
     \label{verification_dnn_desktop_cpu}
\end{figure*}

\begin{figure*}
     \centering
     \includegraphics[width=1.0\linewidth]{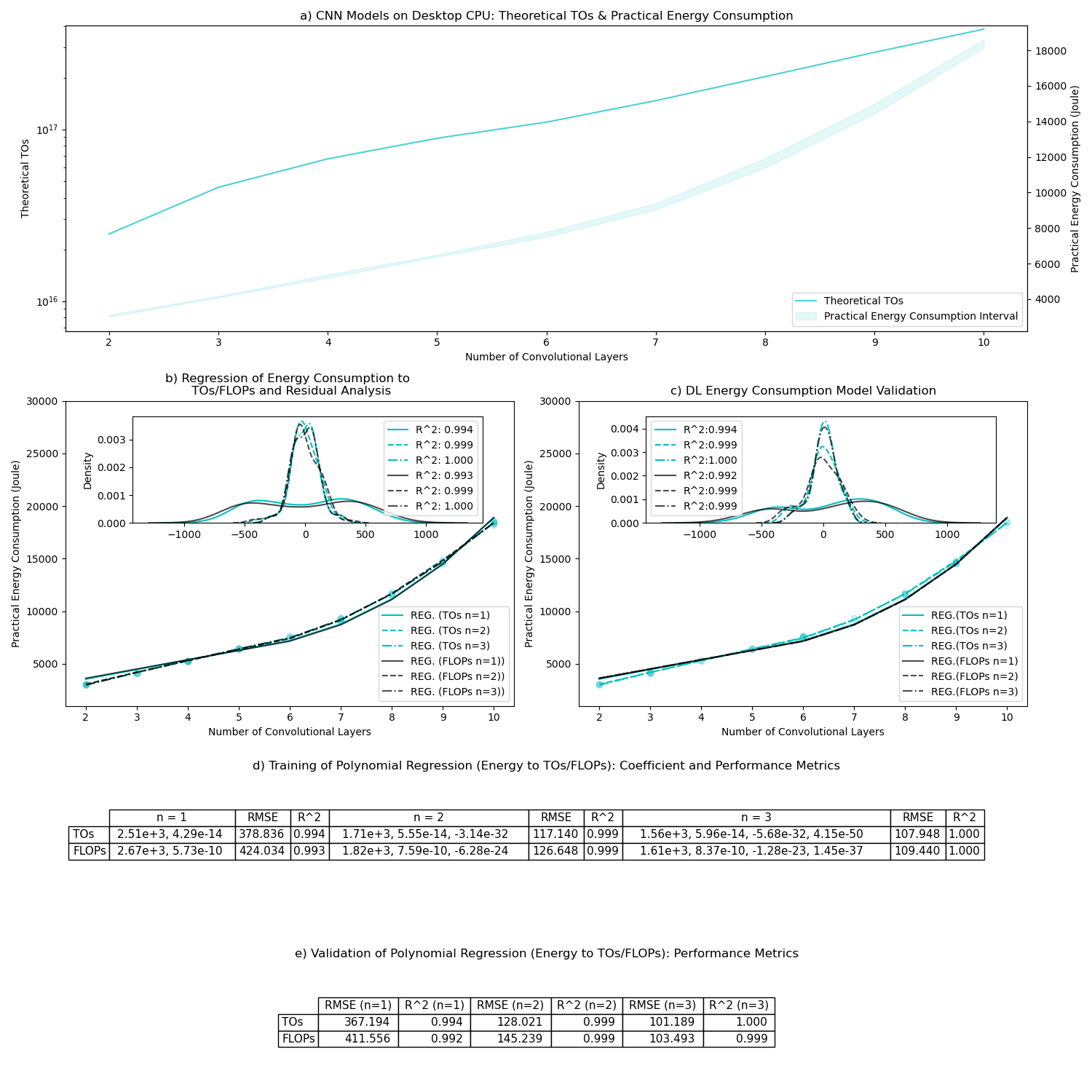}
     \caption{Method Verification (CNN on Desktop CPU)}
     \label{verification_cnn_desktop_cpu}
\end{figure*}

\begin{figure*}
     \centering
     \includegraphics[width=1.0\linewidth]{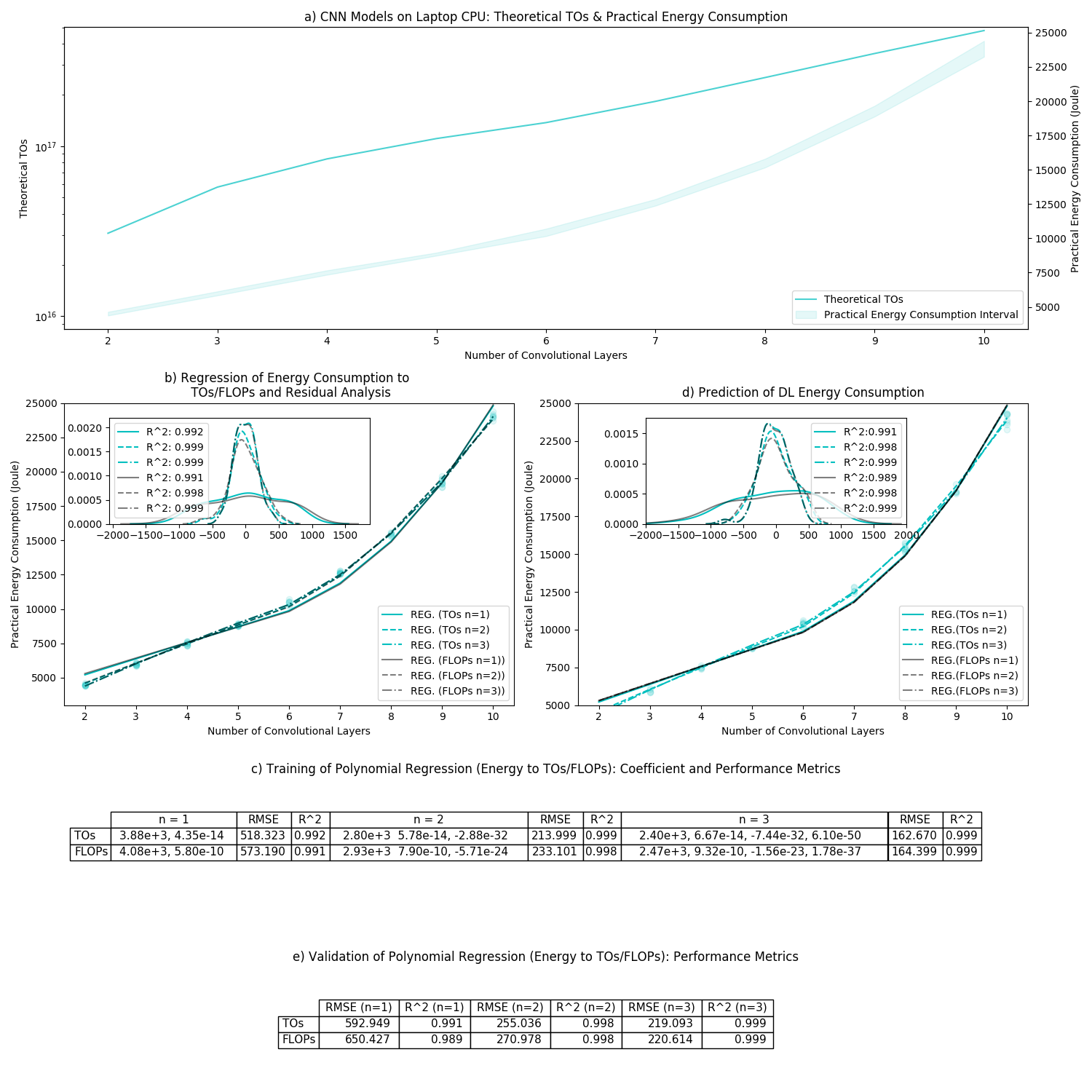}
     \caption{Method Verification (CNN on Laptop CPU)}
     \label{verification_cnn_laptop_cpu}
\end{figure*}
\end{document}